\documentclass{article}

\PassOptionsToPackage{authoryear,round}{natbib}

\usepackage[preprint]{neurips_2026}

\usepackage{amsmath,amssymb,amsfonts,amsthm,mathtools}

\usepackage{graphicx}
\usepackage{booktabs}
\usepackage{nicefrac}
\usepackage{multirow}
\usepackage{subcaption}
\usepackage{tikz}
\usetikzlibrary{arrows.meta,shapes.geometric,positioning}
\usepackage{algorithm}
\usepackage{algorithmic}
\usepackage[table]{xcolor}
\usepackage{listings}

\usepackage[colorlinks=true,linkcolor=blue,citecolor=blue,urlcolor=blue]{hyperref}
\usepackage[capitalize,noabbrev]{cleveref}

\theoremstyle{plain}

\theoremstyle{definition}

\theoremstyle{remark}

\title{Aha-Flow Distillation: Flow Markers Matter in LLM Reasoning}

\author{%
  Xiaodong Wang$^{1,2}$, 
  Peixi Peng$^{1,2\dagger}$ \\
  $^{1}$Peking University, $^{2}$Pengcheng Laboratory \\
  \texttt{\{wangxiaodong21s@stu., pxpeng@\}pku.edu.cn}%
}

\begin{document}

\maketitle

\begin{abstract}
We identify the \textbf{Flow Moment}, a reasoning pattern characterized by
sustained, process-confirming verbalizations such as \emph{I'm doing}, in
contrast to the revision- and backtracking-oriented Aha Moment. We
refer to their corresponding linguistic expressions as Flow Markers
and Aha Markers, respectively. Based on this
observation, we construct Flow-CoT by rewriting the discourse markers of
original reasoning traces while preserving their underlying reasoning content,
and use it as auxiliary supervision for on-policy self-distillation (OPSD).
We further propose \textbf{Aha-Flow Distillation (AFD)}, a dual-mode extension
of OPSD that pairs different forms of privileged information with corresponding
reasoning instructions. The Aha branch retains concise solution-based
supervision, while the Flow branch introduces rewritten Flow-CoT under a
direct and confident reasoning instruction. At inference time, the model uses
only the standard reflective instruction, so Flow-style reasoning serves purely
as a training signal. Experiments on AIME25 and HMMT25 show consistent
improvements across Qwen3-8B and Qwen3-4B: AFD improves Avg@12 from 60.8 to
61.3 on Qwen3-8B and from 57.5 to 58.6 on Qwen3-4B over our reproduced OPSD
baselines. Controlled ablations further show that, with the same
Flow-CoT/Aha-CoT composition, dual-mode training improves Avg@12 from 59.5 to
60.1, indicating that the benefit comes not only from introducing heterogeneous
reasoning supervision, but also from how it is organized during
self-distillation. The code is available at \url{https://github.com/Wang-Xiaodong1899/Aha-Flow-Distillation}.
\end{abstract}

\begin{figure}[h]
    \centering
    \includegraphics[width=\linewidth]{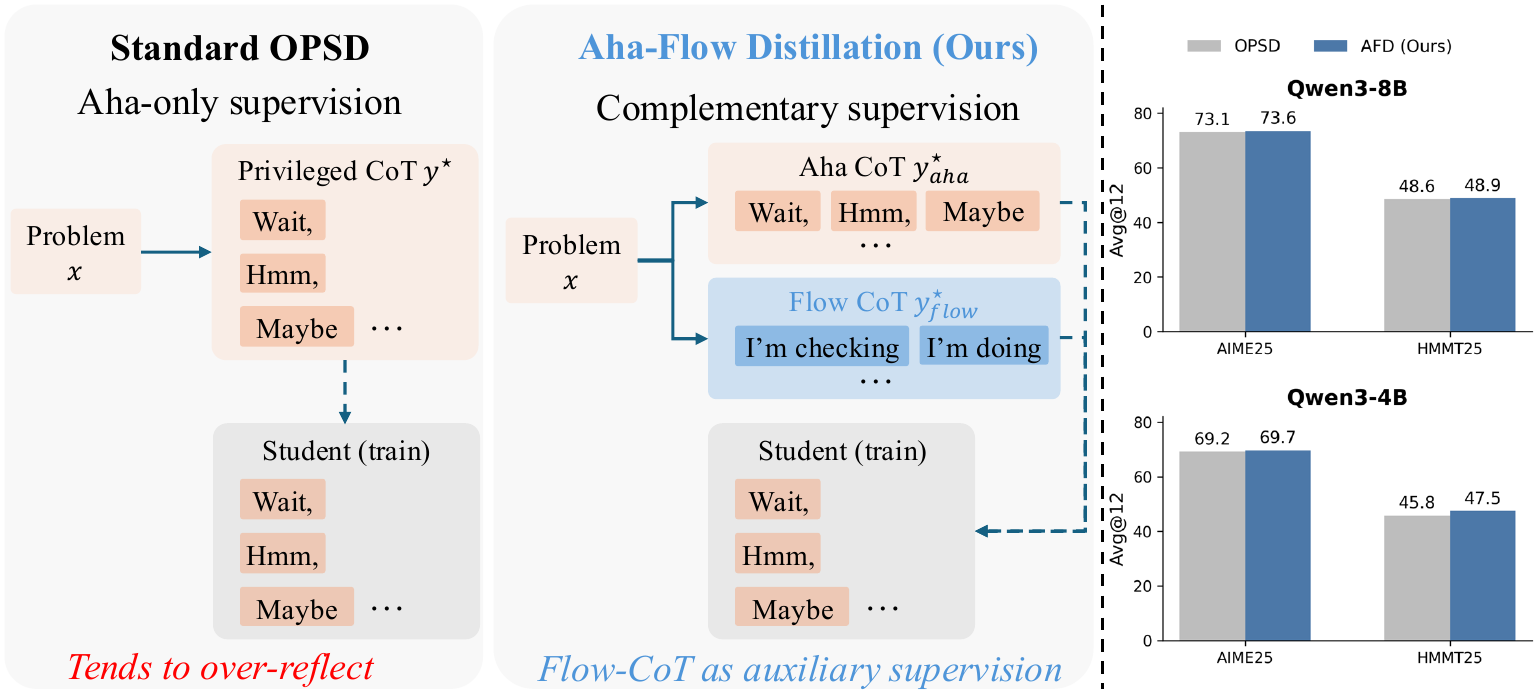}
    \caption{
Comparison between OPSD and our proposed \textbf{Aha-Flow Distillation (AFD)}.
Standard OPSD relies on Aha-style privileged reasoning, while AFD introduces complementary Flow-CoT as auxiliary supervision during training.
The model retains the standard reflective reasoning mode at inference and achieves consistent Avg@12 improvements on AIME25 and HMMT25 across Qwen3-8B and Qwen3-4B.
}
    \label{fig:teaser}
\end{figure}
\section{Introduction}
\label{sec:intro}

Large language models increasingly verbalize their reasoning through long
chain-of-thought (CoT) traces. CoT prompting showed that generating
intermediate reasoning steps can substantially improve multi-step
reasoning~\citep{wei2022chain}, while recent reasoning models such as
DeepSeek-R1~\citep{deepseek2025r1} often exhibit reflection, verification, and
self-correction during reasoning. A characteristic pattern is the
\textbf{Aha Moment}, in which the model reconsiders, backtracks, or redirects
its current reasoning trajectory, often verbalized through expressions such as
\emph{Wait}, \emph{Hmm}, or \emph{Maybe}. Such reflective reasoning has become
prominent in post-trained reasoning models~\citep{xu2026deepseekv4,team2026kimik3,xiao2026mimov2flash}
and is closely associated with training paradigms such as
GRPO~\citep{shao2024deepseekmath} and on-policy
distillation~\citep{agarwal2024onpolicy}. However, excessive reflection can
also introduce redundant reasoning and unnecessary reconsideration of an
otherwise valid solution path. In this work, we study a more confident
reasoning format that proceeds directly along the current trajectory, often
verbalized through expressions such as \emph{I'm doing}. We refer to this reasoning pattern as the
\textbf{Flow Moment} and investigate its role in reasoning-model training.

At the linguistic level, reflective reasoning is often accompanied by short
metacognitive expressions that signal reconsideration or redirection.
Self-correction studies have examined whether such behaviors help models detect
and repair their own mistakes, with mixed evidence on the effectiveness of
intrinsic correction without external feedback~\citep{huang2024cannot,kamoi2024when}.
Related behaviors, including verification, backtracking, and re-evaluation,
also emerge in reasoning models trained with reinforcement
learning~\citep{deepseek2025r1,gandhi2025cognitive}. We refer to expressions
such as \emph{Wait}, \emph{Hmm}, and \emph{Maybe} as \textbf{Aha Markers},
since they typically accompany reflection or a change in the current reasoning
trajectory. In contrast, we refer to expressions such as \emph{I'm doing} as \textbf{Flow Markers}, which
verbalize direct and sustained progression along the current reasoning path.
While Aha Markers have been widely observed and discussed in reflective
reasoning, the role of Flow Markers has received much less attention.

Our interest is in whether the form of reasoning used during post-training
affects the resulting model. To study this question, we focus on on-policy
distillation (OPD)~\citep{agarwal2024onpolicy}, a recent training paradigm in
which supervision is provided on trajectories sampled from the student's own
policy. We further adopt on-policy self-distillation
(OPSD)~\citep{zhao2026opsd}, which removes the need for a separate external
teacher model. In OPSD, the model's teacher forward pass is conditioned on
privileged reference information to supervise its own on-policy generations.
This provides a simple and controlled setting for examining how different
reasoning formats, when used as privileged information, influence
self-distillation.

Based on this setup, we construct Flow-CoT by rewriting a subset of the
original CoT traces, replacing Aha-style discourse markers with Flow Markers
while preserving the underlying reasoning trajectory and mathematical content.
We then propose \textbf{Aha-Flow Distillation (AFD)}, a dual-mode extension of
OPSD. The Aha branch retains the standard step-by-step reasoning instruction
and uses a concise solution as privileged information, while the Flow branch
uses the rewritten Flow-CoT together with an instruction that encourages
direct and confident reasoning. Within each branch, the student and teacher
receive the same reasoning instruction, aligning the teacher supervision with
the behavior requested from the student. At inference time, we use only the
standard reflective instruction. Thus, Flow-CoT serves as auxiliary training
supervision rather than a new inference-time reasoning mode, allowing AFD to
exploit different reasoning formats without introducing additional test-time
procedures.

Experiments with Qwen3-8B and Qwen3-4B on AIME25 and HMMT25 show that the
organization of reasoning styles during distillation matters. In a controlled
ablation using the same 50/50 mixture of Flow-CoT and Aha-CoT as privileged
information, introducing AFD improves the average Avg@12 score from 59.5 to
60.1 on Qwen3-8B, showing that dual-mode training is more effective than
simply mixing heterogeneous CoT references. Our final Flow-CoT/Solution
configuration further reaches 61.3 Avg@12 on Qwen3-8B, compared with 60.8 for
our reproduced OPSD baseline. The same trend holds for Qwen3-4B, where AFD
improves the average from 57.5 to 58.6. AFD also achieves the best AIME25 and
HMMT25 results among the compared methods at both model scales. These results
suggest that the benefit comes not only from introducing diverse reasoning
signals, but also from how different forms of privileged information and
reasoning instructions are paired during self-distillation.

Our contributions are summarized as follows:
\begin{itemize}
    \item We identify the \textbf{Flow Moment} and its corresponding
    \textbf{Flow Markers}, which characterize sustained progression in
    reasoning traces. We further construct Flow-CoT by rewriting the discourse
    markers of original CoT traces while preserving their reasoning content,
    and use it as complementary supervision for self-distillation.

    \item We propose \textbf{Aha-Flow Distillation (AFD)}, a dual-mode
    self-distillation framework that pairs different forms of privileged
    information with corresponding reasoning instructions to better exploit
    heterogeneous reasoning supervision.

    \item Experiments on Qwen3-8B and Qwen3-4B show consistent improvements
    over OPSD on AIME25 and HMMT25. Controlled ablations further verify that
    dual-mode training provides additional gains beyond simply mixing different
    forms of privileged information.
\end{itemize}
\section{Related Work}

\subsection{Reasoning Traces and the Aha Moment}

Chain-of-thought prompting showed that generating intermediate reasoning tokens
can substantially improve multi-step reasoning~\citep{wei2022chain}, even
without few-shot exemplars~\citep{kojima2022zeroshot}. Subsequent work further
treated reasoning traces as a search space, through sampling and aggregation
in self-consistency~\citep{wang2022selfconsistency} or explicit expansion and
pruning in tree-structured search~\citep{yao2023tree}. Across this line of
work, however, reasoning traces are typically treated as largely
undifferentiated sequences, with limited attention to the distinct roles
played by metacognitive expressions within the trace.

One class of such expressions has been extensively studied through
self-correction and reinforcement learning. Intrinsic self-correction is often
unreliable without external feedback~\citep{huang2024cannot,kamoi2024when},
whereas RL with verifiable rewards can induce behaviors such as backtracking
and re-evaluation. DeepSeek-R1 popularized this phenomenon as the ``Aha
Moment''~\citep{deepseek2025r1}, building on scalable RL methods such as
GRPO~\citep{shao2024deepseekmath}; similar behaviors have also been observed
in other long-chain-of-thought systems~\citep{kimiteam2025kimi,yeo2025demystifying}.
Follow-up work suggests that RL amplifies discrete reasoning operations such as
verification, backtracking, subgoal setting, and backward
chaining~\citep{gandhi2025cognitive}. We refer to linguistic expressions
associated with such revision or redirection as Aha Markers.

The utility of Aha-style behavior remains unsettled. Important reasoning steps
are not predominantly self-corrections~\citep{bogdan2025thought}, and weaker
models may switch strategies too frequently and abandon productive
trajectories~\citep{wang2025thoughts}. Our focus is different from determining
whether more Aha behavior is beneficial at inference time. We distinguish Aha Markers from Flow Markers, which verbalize the continuation
of an ongoing reasoning trajectory, and use this distinction to construct an
alternative form of CoT supervision. Specifically, we rewrite Aha-style
discourse markers into Flow-style expressions while preserving the underlying
reasoning trajectory, and use the resulting Flow-CoT as complementary
supervision during self-distillation. To our knowledge, prior work has not
explicitly characterized Flow Markers or investigated Flow-style
verbalization as an auxiliary training signal for reasoning distillation.

\subsection{On-policy distillation for reasoning}

\paragraph{From off-policy to on-policy distillation.}
Knowledge distillation trains a student to imitate a teacher's
outputs~\citep{hinton2015distilling}. For reasoning tasks, the teacher commonly
provides rationales in addition to final labels~\citep{hsieh2023distilling}.
Recent approaches often distill long reasoning traces from stronger models into
weaker ones, either to compress deliberative System-2 computation into more
efficient System-1 behavior~\citep{yu2024distillings2} or to obtain test-time
scaling with smaller models~\citep{muennighoff2025s1}. Such methods are
typically \emph{off-policy}: the student is optimized against traces generated
by another policy, creating a mismatch between training trajectories and the
student's own generation distribution. On-policy formulations reduce this
mismatch by training on student-generated samples that are subsequently scored,
completed, or otherwise supervised by the teacher. This paradigm, commonly
referred to as \emph{on-policy} distillation (OPD), directly distills teacher
supervision on trajectories sampled from the student's current
policy~\citep{agarwal2024onpolicy}. Related approaches include reverse-KL
sequence distillation~\citep{gu2024minillm} and generalized distillation with
mixed teacher/student sampling~\citep{agarwal2024gkd}. More recent variants
extend OPD through entropy-aware objectives~\citep{jin2026eopd},
reward extrapolation beyond the teacher~\citep{yang2026gopd},
black-box teacher supervision~\citep{ye2025black}, and competence-aware
self-distillation focused on the student's learning frontier~\citep{xu2026paced}.

\paragraph{On-Policy Self-distillation.}
On-policy self-distillation (OPSD) removes the need for a separate teacher by
using the model itself to generate on-policy targets while conditioning the
teacher forward pass on a privileged reference solution~\citep{zhao2026opsd}.
This makes the procedure computationally attractive while retaining effective
reasoning supervision, but it also introduces characteristic failure modes.
Prior work reports reduced diversity or freezing of metacognitive markers,
over-confident updates, and, in some settings, degradation of the model's
reasoning capability~\citep{kim2026degrade}. Existing remedies primarily modify
the information content or weighting of the privileged supervision.
Purified OPSD isolates the contamination introduced by the reference solution
and reconstructs the target using pointwise mutual
information~\citep{shen2026purified}, while a parallel line of work modulates
the training update according to the model's own
uncertainty~\citep{ke2026uncertainty}. In contrast, our work focuses on the
reasoning style encoded in self-distillation targets: we distinguish Aha
markers from Flow markers and investigate how their allocation across training
modes affects the resulting policy.

\section{Method}
\label{sec:method}

\begin{figure}
    \centering
    \includegraphics[width=\linewidth]{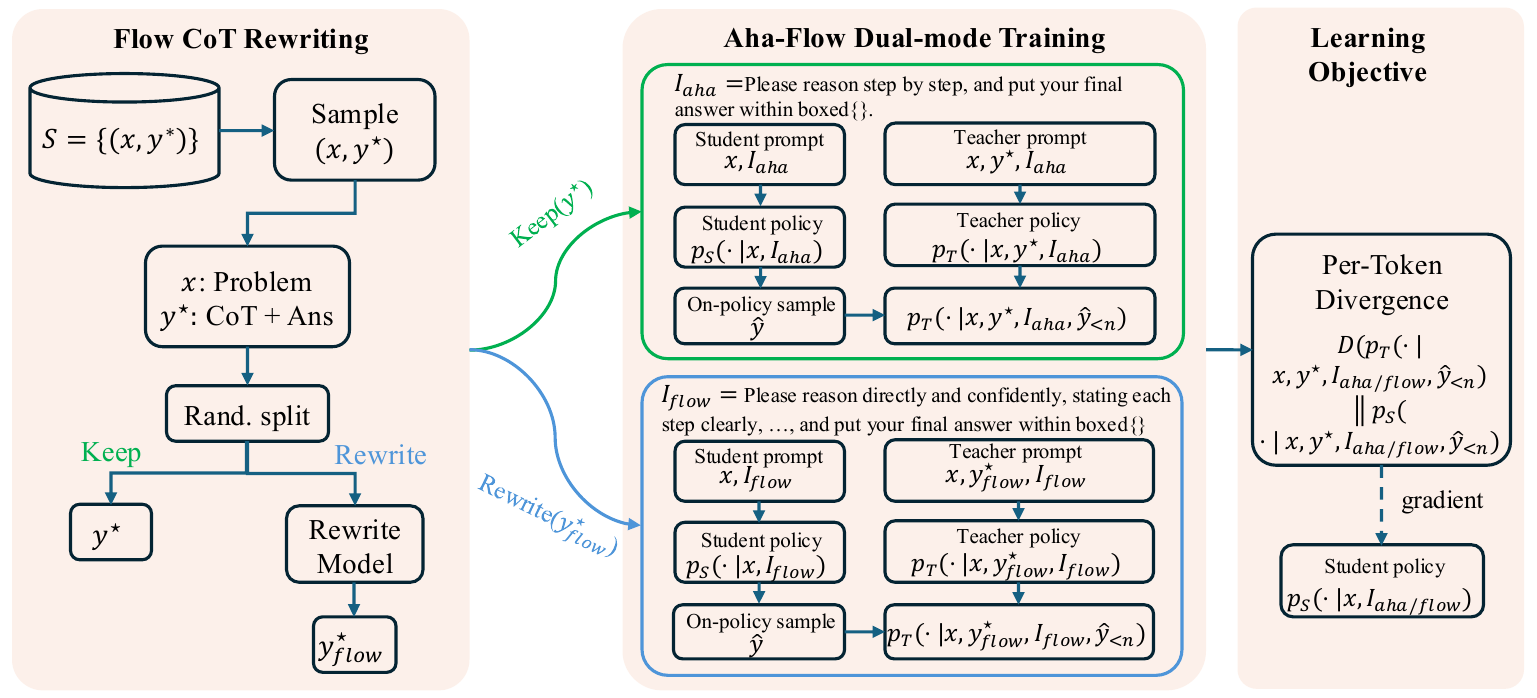}
    \caption{
Overview of \textbf{Aha-Flow Distillation (AFD)}. We randomly rewrite a subset of the original reasoning traces into Flow-CoT while retaining the remaining examples for the Aha branch. AFD then performs dual-mode on-policy self-distillation: the Aha and Flow branches use different privileged information and reasoning instructions, while the student and teacher share the same instruction within each branch.
Both branches are optimized with the token-level JSD objective.}
    \label{fig:main}
\end{figure}

\subsection{Aha Markers and Flow Markers}
\label{sec:taxonomy}

We focus on short metacognitive expressions that indicate how a model manages
its reasoning trajectory. These expressions typically carry little
task-specific content themselves, but signal whether the model is revising or
continuing its current line of reasoning.

We distinguish two types of markers. Let $\mathcal{A}$ denote the set of
\textbf{Aha Markers}, such as \textsc{Wait}, \textsc{Hmm}, and
\textsc{Maybe}. These markers signal a discontinuity in the reasoning
trajectory, including backtracking, re-evaluation, or a change of strategy.
Let $\mathcal{F}$ denote the set of \textbf{Flow Markers}, such as
\textsc{I'm doing} and other process-confirming expressions. These
markers indicate that the current reasoning path is maintained and extended
without redirection.

We use \textbf{Aha Moment} to describe a local reasoning pattern characterized
by revision or redirection, and \textbf{Flow Moment} for its complementary
pattern characterized by sustained progression. Accordingly, Aha Markers and
Flow Markers provide two complementary signals about how a model organizes its
reasoning trajectory.

\subsection{OPSD Preliminaries}
\label{sec:opsd}

On-policy self-distillation (OPSD)~\citep{zhao2026opsd} trains a student on
its own on-policy generations while using privileged reference information to
guide the teacher distribution. Given a training pair $(x,y^\star)$, where
$x$ is the input problem and $y^\star$ is the reference solution, the student
first samples an on-policy response
$\hat{y}\sim p_S(\cdot\mid x)$. At each generation step, the teacher
additionally observes the privileged reference $y^\star$, while both models
are conditioned on the same generated prefix $\hat{y}_{<n}$.

The OPSD objective is as follows:
\begin{align}
\mathcal{L}_{\mathrm{OPSD}}(\theta)
&=
\mathbb{E}_{(x, y^\star) \sim \mathcal{S}}
\;\mathbb{E}_{\hat{y} \sim p_S(\cdot\mid x)}
\sum_{n=1}^{|\hat{y}|} \nonumber\\
&\quad
D\!\Bigl(
p_T\!\left(\cdot \mid x, y^\star,\hat{y}_{<n}\right)
\;\Big\|\;
p_S\!\left(\cdot \mid x, \hat{y}_{<n}\right)
\Bigr),
\label{eq:opsd}
\end{align}
where $p_S$ and $p_T$ denote the student and teacher token distributions,
respectively, and $D(\cdot\|\cdot)$ is the token-level distillation divergence.
Following OPSD, we use Jensen--Shannon divergence (JSD) and clip the per-token
divergence with a threshold $\delta_{\mathrm{clip}}$ to prevent a small number
of tokens from dominating the update~\citep{shen2026purified}.

A key asymmetry is that the teacher has access to $y^\star$, whereas the
student does not. Therefore, the form of the privileged information directly
affects the teacher distribution and, consequently, the supervision received
by the student. This property motivates our investigation of different forms
of privileged information and their associated reasoning styles.

\subsection{Constructing Flow-CoTs}
\label{sec:flow_data}

To construct Flow-style reasoning data, we rewrite the original CoTs that
contain Aha Markers into corresponding Flow-CoTs. We use the
DeepSeek-V4-Flash API~\citep{xu2026deepseekv4} for this transformation. The
rewriting is restricted to discourse-level expressions: Aha-style fillers,
hedges, and transition markers are replaced with explicit first-person action
statements, such as ``I'm checking...'', ``I'm correcting myself...'', or
``I'm trying another approach...''.

Importantly, the underlying reasoning trajectory is kept unchanged. The
rewriting prompt explicitly requires the model to preserve all mathematical
content, intermediate results, dead ends, self-corrections, verification steps,
and their original order, while modifying only the discourse markers. It also
preserves the original paragraph structure and final answer. In this way, each
Flow-CoT remains semantically aligned with its original CoT while differing
primarily in how the reasoning process is verbalized.

Concretely, the rewriting instruction asks the model to convert transition and
hedging expressions into explicit ``I'm $<$verb$>$ing ...'' statements, while
prohibiting any addition, deletion, merging, or reordering of reasoning steps.
The full rewriting prompt is provided in Appendix~\ref{app:rewrite_prompt}.

\subsection{Aha-Flow Distillation}
\label{sec:afd}

We propose \textbf{Aha-Flow Distillation (AFD)}, a dual-mode extension of OPSD
that pairs different forms of privileged information with corresponding
reasoning instructions. The key idea is to make the reasoning behavior
requested from the student consistent with the reference information used to
condition the teacher.

As shown in our ablation study, using a concise solution as privileged
information performs better than using a long CoT under standard OPSD.
We therefore use the solution as the privileged information for the
Aha branch in our final configuration.

For each training example $x$, we define a binary mode indicator
$s(x)\in\{0,1\}$. In the Aha branch, $s(x)=0$, the teacher is conditioned on
the solution $y^\star$. Both the student and
teacher receive the standard reasoning instruction
$\mathcal{I}_{\mathrm{aha}}$:
``\texttt{Please reason step by step, and put your final answer within
\textbackslash boxed\{\}.}''

In the Flow branch, $s(x)=1$, the teacher is conditioned on the rewritten
Flow-CoT reference $y^\star_{\mathrm{flow}}$. Both the student and
teacher receive the Flow-style instruction $\mathcal{I}_{\mathrm{flow}}$:
``\texttt{Please reason directly and confidently, stating each step clearly
without unnecessary hedging, and put your final answer within
\textbackslash boxed\{\}.}''

The privileged reference and instruction for each example are defined as
\begin{equation}
(y^\star(x),\mathcal{I}(x)) =
\begin{cases}
\bigl(y^\star,
      \mathcal{I}_{\mathrm{aha}}\bigr),
      & s(x)=0, \\[2mm]
\bigl(y^\star_{\mathrm{flow}},
      \mathcal{I}_{\mathrm{flow}}\bigr),
      & s(x)=1.
\end{cases}
\label{eq:pairing}
\end{equation}

For each example, the same instruction $\mathcal{I}(x)$ is provided to both the
student and the teacher, while the teacher additionally observes the
privileged reference $y^\star$. This keeps the teacher target aligned with the
reasoning behavior requested from the student and reduces the mismatch between
the demonstrated reasoning style and the student's intended behavior.

This design is motivated by the sensitivity of OPSD to privileged information.
As shown in our experiments, concise solutions generally provide stronger
supervision than long CoT references, while different CoT forms also lead to
different distillation outcomes. AFD therefore does not simply mix
heterogeneous references under a shared instruction. Instead, it pairs each
type of privileged information with a corresponding reasoning instruction:
the Aha branch retains the strong supervision provided by concise
solution-based references, whereas the Flow branch introduces sustained
Flow-style reasoning through both the rewritten CoT and a direct, confident
instruction. In this way, AFD allows the student to benefit from complementary
forms of supervision while preserving the intended role of each branch. The
resulting training signal is both more diverse and better aligned, enabling
heterogeneous reasoning behaviors to be incorporated into self-distillation
more effectively.

\section{Experiments}
\label{sec:exp}

\subsection{Experimental Setup}
\label{sec:setup}

\paragraph{Training setup.}
We use Qwen3-8B and Qwen3-4B~\citep{yang2025qwen3} as backbone models and follow the OPSD training
recipe~\citep{zhao2026opsd}. The student model is adapted with LoRA using rank
$r=64$ and scaling factor $\alpha=128$, applied to all projection layers.
During the teacher forward pass, the LoRA adapters are disabled, keeping the
teacher fixed throughout training. By default, thinking mode is disabled for the student and enabled for the teacher. The maximum training completion length is set to 1,024 tokens.

Following~\citep{zhao2026opsd}, we use OpenThoughts-30K~\citep{guha2026openthoughts}
as the base training dataset. We construct the Flow-CoT data by using the
DeepSeek-V4-Flash API~\citep{xu2026deepseekv4} to rewrite the original
Aha-style CoT traces while preserving their underlying reasoning content.
The complete rewriting prompt is provided in Appendix~\ref{app:rewrite_prompt}.

\paragraph{Evaluation protocol.}
We evaluate on two mathematical-reasoning benchmarks, AIME25 and HMMT25.
For each problem, we sample 12 completions using vLLM with tensor-parallel
size 8, temperature $1.0$, thinking mode enabled, and a maximum generation
length of $38\mathrm{k}$ tokens. We report $\mathrm{Avg@12}$, computed as
the mean correctness over the 12 sampled completions and then averaged across
all problems. All models are evaluated with exactly the same generic reflective instruction
and decoding configuration. 
\subsection{Main Results}
\label{sec:main}

\begin{table*}[t]
\centering
\caption{
Main results on mathematical-reasoning benchmarks.
All entries are $\mathrm{Avg@12}$.
Evaluation follows the Qwen3 sampling configuration with temperature $1.0$
and a maximum generation length of $38\mathrm{k}$ tokens.
${^\dagger}$ denotes our reproduction.
The best result in each model group is shown in bold.
}
\label{tab:main_results}
\small
\setlength{\tabcolsep}{12pt}
\begin{tabular}{l|ccc}
\toprule
\textbf{Method}
& \textbf{AIME25}
& \textbf{HMMT25}
& \textbf{Average} \\
\midrule
\multicolumn{4}{l}{\textit{Qwen3-8B}} \\
\quad Base (Instruct)
& 65.6 & 43.9 & 54.8 \\
\quad + SFT
& 64.2 & 42.9 & 53.6 \\
\quad + GRPO~\citep{deepseek2025r1}
& 68.9 & 46.7 & 57.8 \\
\quad + OPSD~\citep{zhao2026opsd}
& 70.8 & 45.8 & 58.3 \\
\quad + OPSD${^\dagger}$
& 73.1 & 48.6 & 60.8 \\
\rowcolor{gray!15}
\quad + AFD (Ours)
& \textbf{73.6} & \textbf{48.9} & \textbf{61.3} \\
\midrule
\multicolumn{4}{l}{\textit{Qwen3-4B}} \\
\quad Base (Instruct)
& 66.4 & 42.2 & 54.3 \\
\quad + SFT
& 62.3 & 43.4 & 52.8  \\
\quad + GRPO~\citep{deepseek2025r1}
& 68.1 & 44.4 & 56.3 \\
\quad + OPSD~\citep{zhao2026opsd}
& 68.3 & 46.1 &  57.2 \\
\quad + OPSD${^\dagger}$
& 69.2 & 45.8 & 57.5 \\
\rowcolor{gray!15}
\quad + AFD (Ours)
& \textbf{69.7} & \textbf{47.5} & \textbf{58.6} \\
\bottomrule
\end{tabular}
\end{table*}

As shown in \cref{tab:main_results}, AFD consistently improves over our
reproduced OPSD baseline across both model scales and both benchmarks.
On Qwen3-8B, AFD achieves 73.6 on AIME25 and 48.9 on HMMT25, compared with
73.1 and 48.6 for OPSD${^\dagger}$, increasing the average from 60.8 to 61.3.
The improvement is larger on Qwen3-4B: AFD reaches 69.7 on AIME25 and 47.5 on
HMMT25, compared with 69.2 and 45.8 for OPSD${^\dagger}$, raising the average
from 57.5 to 58.6.

These results show a consistent numerical advantage of AFD across model
scales. In particular, the improvement over OPSD${^\dagger}$ is observed on
both benchmarks for both Qwen3-8B and Qwen3-4B, suggesting that the effect is
not specific to a single benchmark or model size. Since all models use the
same reflective prompt at evaluation, the gains arise from the training
procedure rather than additional test-time prompting.

\subsection{Ablation Study}
\label{sec:ablation}

\begin{table*}[t]
\centering
\caption{
Ablation of privileged information and dual-mode training on Qwen3-8B.
Rows containing two forms of privileged information use an equal 50/50
mixture. All other training and evaluation settings are held fixed.
The best result in each column is shown in bold.
}
\label{tab:abl}
\small
\setlength{\tabcolsep}{9pt}
\begin{tabular}{lc|ccc}
\toprule
\textbf{Privileged information}
& \textbf{Training mode}
& \textbf{AIME25}
& \textbf{HMMT25}
& \textbf{Average} \\
\midrule
\multicolumn{5}{l}{\textit{Qwen3-8B}} \\
\quad --
& --
& 65.6 & 43.9 & 54.8 \\
\midrule
\quad Solution
& OPSD
& 73.1 & 48.6 & 60.8 \\
\quad Aha-CoT
& OPSD
& 70.8 & 47.5 & 59.2 \\
\quad Flow-CoT / Aha-CoT
& OPSD
& 70.8 & 48.1 & 59.5 \\
\quad Flow-CoT / Aha-CoT
& AFD
& 72.8 & 48.3 & 60.1 \\
\rowcolor{gray!15}
\quad Flow-CoT / Solution
& AFD
& \textbf{73.6} & \textbf{48.9} & \textbf{61.3} \\
\bottomrule
\end{tabular}
\end{table*}

\cref{tab:abl} first shows that the form of privileged information has a
substantial effect on OPSD. Using the solution alone gives an average
score of 60.8, outperforming Aha-CoT, which obtains 59.2. The difference is
observed on both benchmarks: solution-based OPSD reaches 73.1 on AIME25 and
48.6 on HMMT25, whereas Aha-CoT reaches 70.8 and 47.5, respectively. Mixing
Flow-CoT and Aha-CoT under standard OPSD slightly improves the average over
Aha-CoT alone, from 59.2 to 59.5, but remains below the solution-based
configuration. These results suggest that, as privileged information, a more
direct and concise solution is generally more effective than a lengthy
chain-of-thought trace. Beyond this general comparison, we are particularly
interested in whether different forms of CoT supervision lead to different
distillation outcomes.

We isolate the effect of dual-mode training by comparing the two
Flow-CoT/Aha-CoT configurations. They use the same 50/50 composition of
privileged information and differ only in training mode. Standard OPSD obtains
70.8 on AIME25 and 48.1 on HMMT25, with an average of 59.5. Enabling AFD
raises these scores to 72.8, 48.3, and 60.1, respectively. Because the
privileged-information mixture is unchanged, this comparison indicates that
the improvement cannot be explained by target composition alone. Explicitly
pairing heterogeneous privileged information with corresponding reasoning
instructions provides a more effective distillation signal than presenting
the same mixture under a single training mode.

Finally, the Flow-CoT/Solution configuration achieves the strongest overall
performance, reaching 73.6 on AIME25, 48.9 on HMMT25, and 61.3 on average.
Compared with Flow-CoT/Aha-CoT under AFD, replacing Aha-CoT with the refined
solution improves both benchmarks and raises the average from 60.1 to 61.3.
It also improves over solution-only OPSD, whose average is 60.8. Together,
these results suggest two complementary conclusions: the choice of privileged
information materially affects self-distillation, and dual-mode training
provides additional benefit beyond simply mixing different forms of privileged
information.

\section{Conclusion}
\label{sec:conclusion}

We identify the \textbf{Flow Moment}, a reasoning verbalization pattern that
signals sustained progression, as a complementary counterpart to the
correction-oriented Aha Moment. Based on this observation, we construct
Flow-CoT by rewriting discourse markers while preserving the underlying
reasoning trajectory, and use it as auxiliary supervision for on-policy
self-distillation. We further propose \textbf{Aha-Flow Distillation (AFD)},
which pairs different forms of privileged information with corresponding
reasoning instructions. Experiments on Qwen3-8B and Qwen3-4B show consistent
improvements over OPSD on AIME25 and HMMT25, while the model continues to use
the standard reflective reasoning mode at inference. More broadly, our results suggest that the form and organization of privileged
reasoning information are important design choices for self-distillation,
motivating further study of alternative reasoning verbalizations and their use
as auxiliary supervision.

\bibliography{references}

\newpage
\appendix

\section{Flow-CoT Rewriting Prompt}
\label{app:rewrite_prompt}

We use DeepSeek-V4-Flash~\citep{xu2026deepseekv4} to rewrite the original
reasoning traces into Flow-CoT. The rewriting is restricted to discourse
markers and is designed to preserve the original reasoning trajectory and
mathematical content as closely as possible. The complete system prompt and
user template are shown below.

\begin{lstlisting}[basicstyle=\ttfamily\footnotesize,breaklines=true]
SYSTEM_PROMPT = """You are a precise text editor. You rewrite the *discourse markers* of a mathematical reasoning transcript without touching its mathematical content.

Rewrite rules:
1. Replace every filler, hedging, or transition marker with an explicit first-person action statement in the "I'm <verb>ing ..." form. Examples:
   - "Okay, let's try to tackle this problem step by step." -> "I'm tackling this problem step by step."
   - "Hmm, this seems complex." -> "I'm noticing that this is getting complex."
   - "Wait, actually the formula is ..." -> "I'm correcting myself: the formula is ..."
   - "But wait, let me check the constraints." -> "I'm now checking the constraints."
   - "Alternatively, maybe using symmetric sums." -> "I'm trying another approach: symmetric sums."
   - "Not directly obvious. Maybe not. Let's try another approach." -> "I'm finding no direct correspondence, so I'm dropping this idea and trying another approach."
   - "Let me compute each term." / "Compute each term:" -> "I'm computing each term:"

2. Preserve the reasoning trajectory exactly: keep every dead end, self-correction and verification step. Do NOT solve the problem yourself. Do NOT add, remove, merge or reorder any reasoning step.

3. Preserve ALL mathematics verbatim: every formula, variable, number, intermediate result, LaTeX command and unicode symbol (lambda, x_1, ^3, sqrt, Delta, ...).

4. Preserve the paragraph and blank-line structure, and keep the trailing "**Final Answer**" line together with its \boxed{...} exactly as-is.

5. Output ONLY the rewritten transcript. No preamble, no commentary, no code fences, no explanation of what you changed."""
\end{lstlisting}

The corresponding user template is:

\begin{lstlisting}[basicstyle=\ttfamily\footnotesize,breaklines=true]
USER_TEMPLATE = """Rewrite the discourse markers of the transcript below according to the rules. Return the full rewritten transcript only.

=== TRANSCRIPT BEGIN ===
{text}
=== TRANSCRIPT END ==="""
\end{lstlisting}

The prompt is intentionally conservative: it instructs the rewriting model to
change only the verbalization of the reasoning process while preserving the
sequence of reasoning steps, including unsuccessful attempts, self-corrections,
and verification. Therefore, the resulting Flow-CoT differs from the original
CoT primarily in its discourse markers rather than its mathematical reasoning
content.

\end{document}